\documentclass[a4paper, 11pt]{article}

\usepackage{hyperref}
\usepackage{pgf}
\usepackage{amsmath,amssymb,amsfonts}
\usepackage{natbib}
\usepackage{xcolor}
\definecolor{lnkcol}{rgb}{0,0,0.93}

\hypersetup{
    pdfauthor={Hugo Martin, Sylvain Chevallier and Eric Monacelli},     
    pdfsubject={Adaptative Visualisation System for Construction Building Information Models Using Saliency},   
    pdfnewwindow=true,      
    pdfkeywords={Saliency, Building Information Modeling, Virtual Reality, Building Architecture, Data Visualization, Immersion, Knowledge Representation}, 
    colorlinks=true,        
    linkcolor=lnkcol,       
    citecolor=lnkcol,       
}

\begin{document}

\title{Adaptive Visualisation System for Construction Building Information Models Using Saliency}

\author{Hugo Martin, \texttt{hug.martin@bouygues-construction.com}\\
  Sylvain Chevallier, \texttt{sylvain.chevallier@uvsq.fr}\\
  Eric Monacelli, \texttt{eric.monacelli@uvsq.fr}}



\maketitle

\begin{abstract}

Building Information Modeling (BIM) is a recent construction process based on a 3D model, containing every component related to the building achievement. 
Architects, structure engineers, method engineers, and others participant to the building process work on this model through the design-to-construction cycle.
The high complexity and the large amount of information included in these models raise several issues, delaying its wide adoption in the industrial world.
One of the most important is the visualization: professionals have difficulties to find out the relevant information for their job. 
Actual solutions suffer from two limitations: the BIM models information are processed manually and insignificant information are simply hidden, leading to inconsistencies in the building model.
This paper describes a system relying on an ontological representation of the building information to label automatically the building elements.
Depending on the user's department, the visualization is modified according to these labels by automatically adjusting the colors and image properties based on a saliency model.
The proposed saliency model incorporates several adaptations to fit the specificities of architectural images.

\end{abstract}




\section{Introduction}

\begin{figure}[ht]
\centering
\includegraphics[width=1\linewidth]{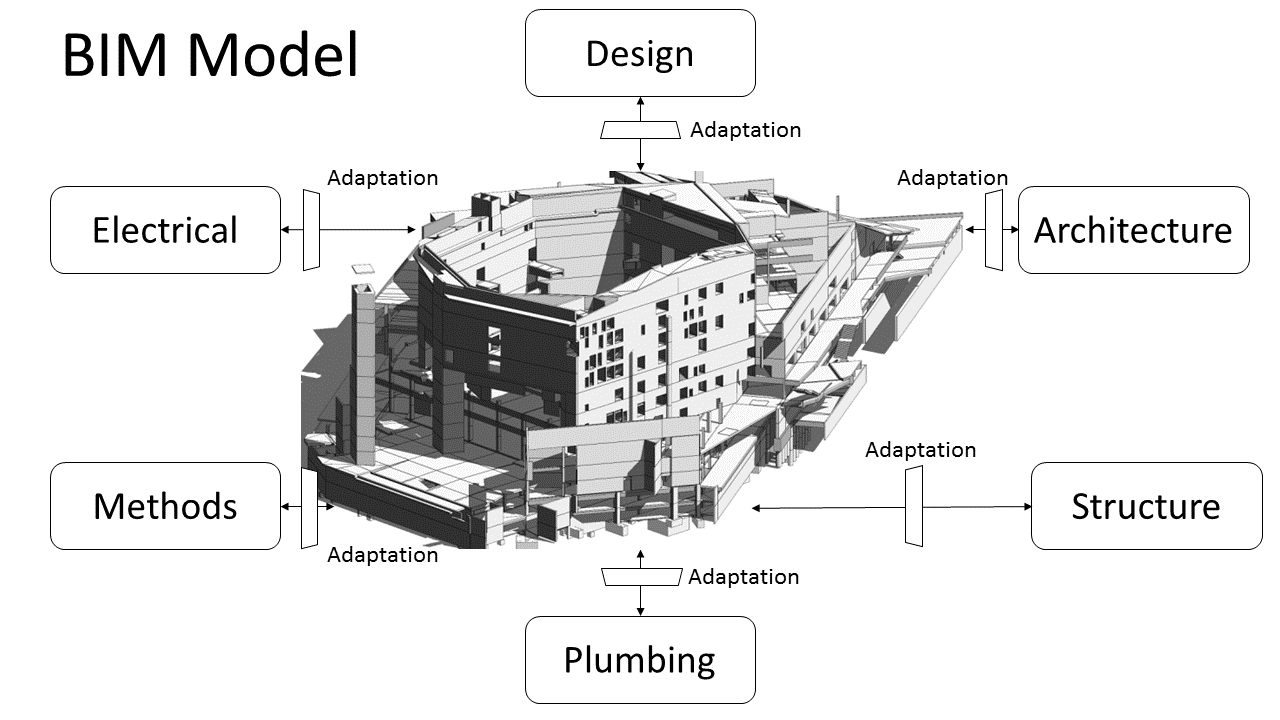}
\caption{Sketch of BIM process. All professions work on the same model. Each user needs an adaptation of his visualisation of the model.}
\label{fig:BIM}
\end{figure}

In architecture, the construction process underwent a major shift in the past 5 years with the increasing adoption of Building Information Modeling process (BIM) to design and achieve all kind of structures.
These processes are based on a 3D model containing the information required to achieve the building.
The model could be used to observe the construction project in term of design and architecture with lightings simulations or rendering pictures, but it is mostly used in a technical approach~\citep{ERN06}~\citep{CRC07}~\cite{KIM08}. 
Several distinct experts are contributing to BIM models, each one modifying or adding its own objects and notes on the 3D model. 
As shown in Fig.~\ref{fig:BIM}, this 3D model gathered all the information coming from different user profiles (architect, method, structure, $\ldots$).

All user profiles do not require all information from the model; e.g. the position of electric cables is mostly irrelevant for a structure engineer, but it is needed at some point to check interactions between electric cables and structures.
To improve the efficiency and quality of the BIM work, it is important to adapt the visibility of each object depending of the user's profile. 
The actual industrial process is to manually hide with a transparency effect all the unimportant objects. 
This method is the cause of major problems, as the manual operations are known to be error-prone and time consuming.
Also completely hiding objects often lead to clashes between elements, for example a cable going through a water pipe, which is a common issue in BIM processes.

This paper proposes an assistive system for construction engineers and BIM users. 
This system uses an ontology to represent the building information and enhance the identification of 3D models. 
It classifies automatically BIM components in real time and adapts their colors regarding the user's interest. 
Our system proposes a visualization method where the most important components capture the visual attention while maintaining all components on screen.

An eye tracking device could be used to accurately capture the gaze of the user during their interaction with the system, however this approach suffers from two limitations.
Firstly, the design and parametrization of a system based on eye-tracking recordings is a time-consuming process.
It implicates that building experts spend a lot of time calibrating the system, thus the effective cost will be very high.
Secondly, with eye-tracking device it is not possible to produce an online system, i.e. which automatically adapts the visualization depending on the properties of the visual scene.
For example, under a certain point of view, an object could require an enhancement to be salient, while it could already be salient when looking at it from another point of view.

Saliency models are mostly used for image analysis and capturing the focus of human perception.
In this work, we make use of saliency model for image synthesis and we exploit the efficiency of these models to propose an online system.
The visualization of the 3D model is thus modified to enforce the saliency of the relevant objects for a specific user profile.

Most of the saliency models are assessed on natural images, such as landscapes, portraits, etc. 
The architectural images generated by 3D building models display some strong properties and different from natural images.
The most visible property is the strong presence of vanishing points and the second one is the importance of depth in those images. 
We proposed specific adaptations of the saliency models to suit those properties of architectural images.
This work will then be integrated in an industrial BIM viewer for Bouygues Construction and applied on the design of real projects. 

Thus, the contributions of our system could be summarized as:
\begin{itemize}
\item Ontology of building information models: from the knowledge of BIM managers, we propose an ontology and a classification of BIM elements depending of their interest for each profession department.
\item Adaptation of the saliency model to architectural images: Notable saliency systems are mostly based on 2D images. BIM and architectural images present particularities that we take into account. 
\item Online and real-time visualization system: We propose a system which modify colors depending of the user profile. The system works in real time and is integrated in a BIM viewer to assist users.
\end{itemize}

This paper is organized as follows.
Section~\ref{sec:soa} describes the state of the art on building management ontology and defines our goals for the system. 
The application of saliency methods are also justified and the specificities of architectural images are pointed out.
Section~\ref{sec:model} introduces our model and explain our contributions. 
Section~\ref{sec:expe} details the experiments and in Sect.~\ref{sec:ccl} a discussion about benefits of the proposed system is proposed.

\section{State of the Art}
\label{sec:soa}
\subsection{Building Management Ontology}

Two different approaches could be identified in the literature on ontology applied to building management.
The first approach is task-centric and aims at assessing the benefit of BIM versus non-BIM technologies, relying on a complete description of the tasks taking place in the building design process~\cite{LI14}. 
The work developed in this paper follows a second approach, user profile-centric : all 3D model objects are classified depending on their relation with the user’s profile.
This classification is either based on the intrinsic properties of the object~\cite{AKS07} or from a model of interaction between different actors~\cite{LEE14}~\cite{COS14}.
As such, an element of the 3D model, for example a wall, is relevant for the work of a structure engineer, but is not for a plumbing engineer who is concerned by pipes and water circulation.



\subsection{Saliency}

Many computational models have been proposed to predict the visual attention through saliency map, such as~\cite{TSO95}~\cite{ITT00}~\cite{HUA12}.
A comprehensive review could be found in~\cite{ZHA13} and an assessment of the existing models is proposed on the MIT saliency benchmark~\cite{mit-saliency-benchmark}.
A commonly found methodology in saliency-based models is to analyze the contrast of an image in term of intensity, color and orientation. 
Saliency models are an easy and fast way of analyzing images and predicting an accurate map of visual attention.
Hence saliency models could analyze an image coming from a BIM scene and determine the visually attractive elements of the scene. 

In the context of architectural images, the visual scene often display strong depth information and vanishing point.
It has been shown that the depth clues are highly informative and have a strong influence on object saliency~\cite{LAN12,HE95}. 
Several works linking vanishing point and visual attention have been conducted. 
Stentiford~\cite{STE06} has proposed to detect vanishing points based on the visual attention. 
Itti and Koch model integrated the orientation analysis following constant axes (vertical, horizontal, diagonal). 
Architecture pictures contain important vanishing points which are easily detectables~\cite{HEU98,ROT02} as most of the shapes follow the projection of vanishing points.
If an object does not contribute to the global perspective, its visual saliency is increased. 

\section{Ontology-based Visualization System}
\label{sec:model}



\subsection{Description of the Proposed System}

The majority of the saliency models are developed using standard 2D pictures. 
These models consider orientations in flat images, the visual saliency is computed for all constant angles. 
Our system implement a method that detects vanishing points and compute orientation map from a projection of these points, described in Sect.~\ref{sec:vp}.
In BIM case, all images are generated by a 3D viewer software with engineer colors only chosen to make a contrast between elements. 
The employed color codes do not contain major information; often colors are chosen randomly. 
We used saliency model for image synthesis, guiding the color choice to maximize the saliency of important BIM components in the scene.
Combined to a color selection algorithm, our system selects colors which keep the visual attention of the user on important elements without hiding or deleting others.
Our algorithm is described in Sect.~\ref{sec:coloradapt}.


\subsection{Vanishing Point-Sensitive Computation}
\label{sec:vp}

\begin{figure}[t]
\centering
\includegraphics[width=1\linewidth]{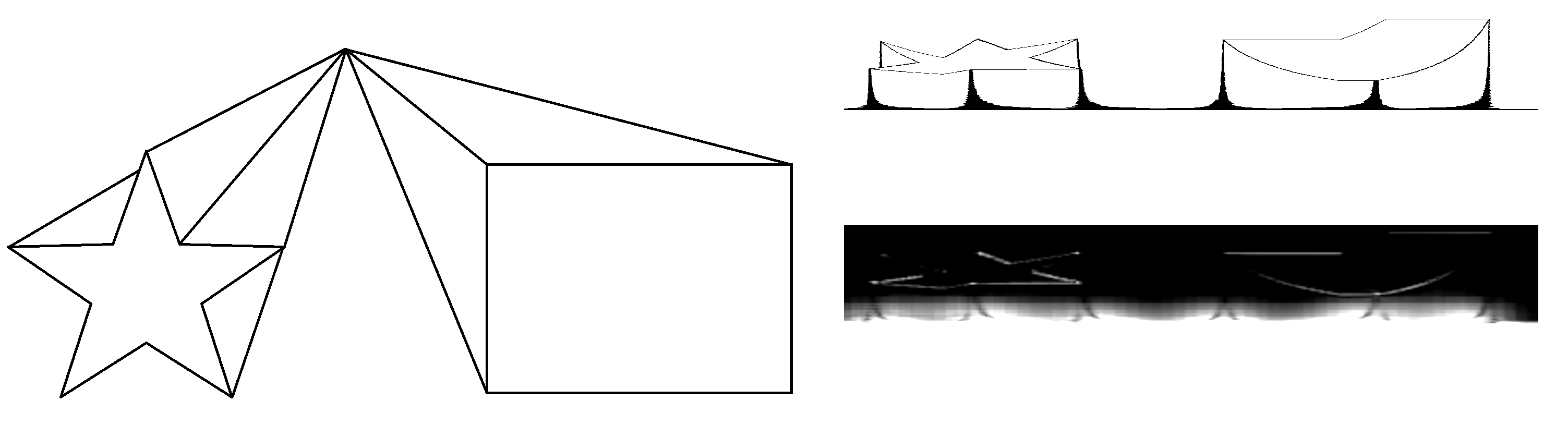}
\caption{Left: original image with obvious vanishing point. Top right: result image after vanishing point projection. Bottom right: the vertical orientation map. }
\label{fig:projection}
\end{figure}

\begin{figure}[t]
\centering
\includegraphics[width=1\linewidth]{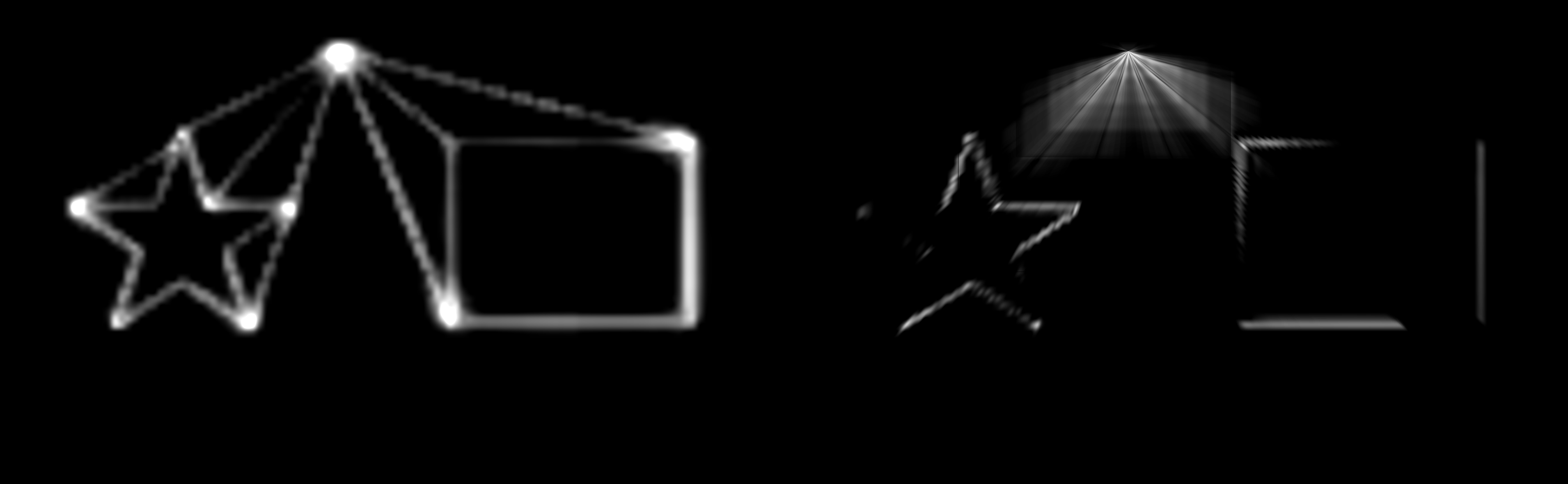}
\caption{Left: orientation conspicuity map computed from ${0^°, 45, 90, 135}$ orientations, as in Itti \emph{et al.}~\cite{ITT00}. Right: orientation conspicuity map computed with our model. Edges that do not contribute to the perspectivity stood out. }
\label{fig:comparaison}
\end{figure}

This paper proposes a method  to take into account vanishing point in saliency model. 
The orientation detection was replaced by a vanishing point projection process.
Our system relies on the work of Feng \emph{et al.} for the detection of the vanishing points~\cite{FEN10}~\cite{TAR09}.
As an illustration, we applied our algorithm on a sample image and the result is shown Fig.~\ref{fig:projection}.
From an input image, the vanishing point are detected with the Feng algorithm, the projection from this vanishing point is computed.
The projection image (Fig.~\ref{fig:projection}, top right) is then analyzed in its vertical contrast (Fig.~\ref{fig:projection}, bottom right). 
An inverse transform of the result is added to the global model as the orientation conspicuity map, as shown on the right part of Fig.~\ref{fig:comparaison}. 
This figure displays the saliencies computed with the original Itti model~\cite{ITT00} on the left and the saliencies obtained with our model on the right.

\begin{figure}[ht]
\centering
\includegraphics[width=1\linewidth]{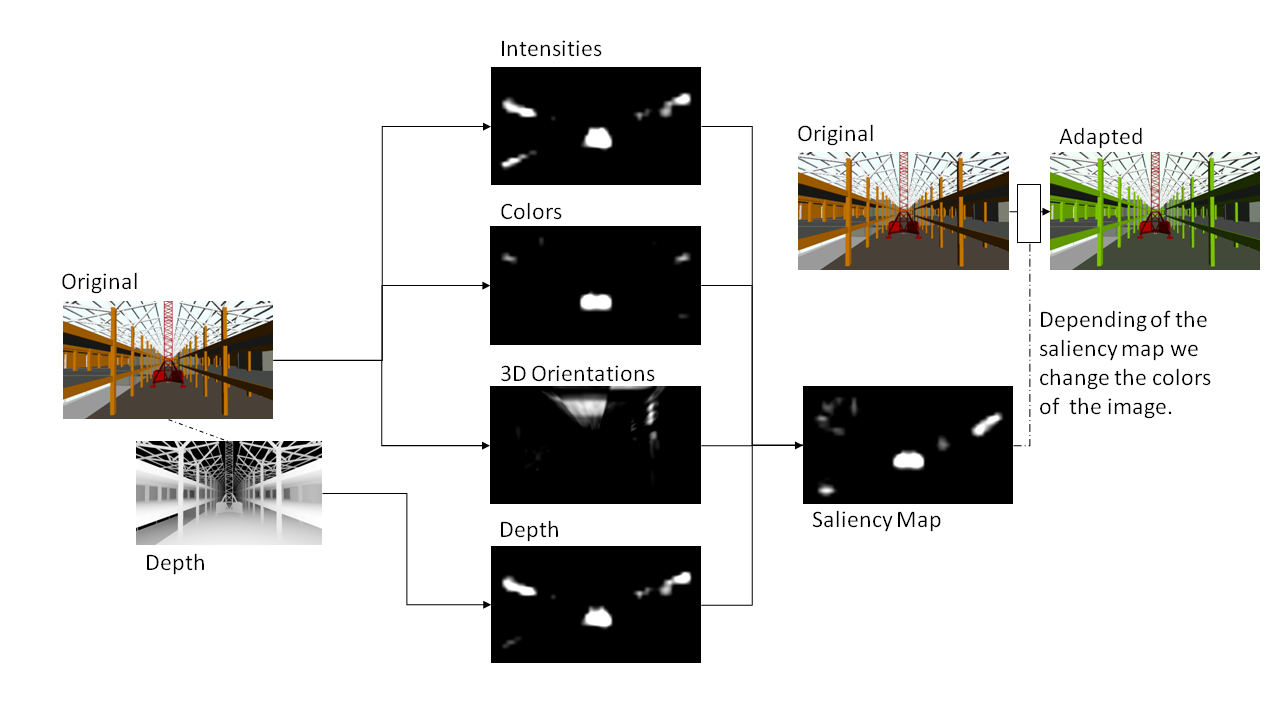}
\caption{Architecture of the saliency model: conspicuity maps of intensity, colors and 3D orientations are extracted from the 3D snapshot. Depth conspicuity map is computed from depth information of the model. The saliency map combines all the conspicuity maps. Salient locations are used to adapt the color scheme of the original image to enhance object relevant to the user profile.}
\label{fig:final}
\end{figure}

The saliency model of the proposed system takes also into account the contrast of depth.
The 3D software computes a depth map associated with each generated image.
We have implemented the work of Jost \emph{et al.}~\cite{JOS04} to include the information extracted from a depth conspicuity map in the system.
In the proposed saliency model, other features map are computed as in the Itti model~\cite{ITT00}.
Thus four conspicuity maps are built, i.e. the perspective orientation map, the intensity map, the color map and the depth map, and are then combined in the saliency map, shown on Fig.~\ref{fig:final}.

\subsection{Enhancing Visual Attention}
\label{sec:coloradapt}

The proposed system implements an online adaptation the image visualization.
The aim is to choose colors that emphasize the visual saliency of the relevant elements. 
The works of Ware~\cite{WAR04} and Healey~\cite{HEA12} describe links between visual attention, visualization and colors. 
The thorough experiments of Wolfe~\cite{WOL00} caracterize the links between colors and visual attention. 
Colors of BIM model do not convey information, and are chosen semi-randomly only to create a difference between several elements. 
In the proposed approach, most of the adaptation of the original image is done on the color channels, as it is a lossless transformation regarding the BIM model.
It should be noted that as BIM models become more complex and include more elements, the probability of finding two different objects with a similar color  drastically increase, thus producing incomprehensible visualization of BIM models. 
Our system does not alter the user-defined colors, but could modify the color of any element label as unimportant with respect to the user profile.

As shown on Fig.~\ref{fig:final}, the saliencies are computed from a screenshot provided by the 3D real time viewer. 
From the obtained saliency map, regions belonging to relevant and irrelevant elements are tagged as such. 
$\mathcal{R}_r = \{p_a\}_{a \in A}$ and $\mathcal{R}_i = \{p_b\}_{b \in B}$ are respectively the set of relevant and irrelevant regions in the image, where $A$ and $B$ are the indexed set of pixel $p_a$ and $p_b$ belonging to relevant and irrelevant objects. 
Sky and gound's pixel are considered as irrelevant objects.
The average saliency $\mathcal{S}_r$ and $\mathcal{S}_i$ of $\mathcal{R}_r$ and $\mathcal{R}_i$ regions is computed to evaluate which set is the most salient. 
\begin{align}
  \label{eq:Pe}
  G_e & = \frac{\mathcal{S}_r}{\mathcal{S}_r+\mathcal{S}_i}\\
  Si & = {p_a \cap p_i : a \in A, i \in I}\\
\end{align}

The colors are altered if $G_e< 0,5$, that is if the $\mathcal{R}_r$ region does not represent the majority of the image saliency.
The color modifications rely on the average saturation, intensity and color (in the HSV sense), denoted $G_e$, measured on $\mathcal{R}_r$. 
Depending on the average element color, a target color $T$ is defined (see Table~\ref{tab:target}). 
This target color is the opposite of the $\mathcal{R}_r$ pixel color, as the main idea is to reinforce the color contrast of the image depending on the relevant image elements.


\begin{table}[ht]
\centering
\begin{tabular}{| l | c |}
 \hline
 Average Color & Target ($T$) \\
 \hline
Red & Green \\ 
  \hline
Yellow & Blue \\ 
  \hline
Green & Red \\ 
  \hline
Blue & Yellow \\ 
\hline
\end{tabular}
\caption{Target color $T$ of the algorithm modifying the visualization to enhance the saliency of relevant element w.r.t the user profile.}
\label{tab:target}
\end{table}

The proposed algorithm applies a color filter to turn the less important pixels depending of the user's profile ($p_b \subset \mathcal{R}_i$) in the target hue, resulting in a new pixel $\tilde{p}_b$. 
The same filter is applied to the black and white pixels.

\begin{equation}
\tilde{p}_b = p_b + G_e (T - p_b)
\label{eq:2}
\end{equation}

A new saliency map is computed with the updated color map.
If the average saliency of the $\mathcal{R}_r$ failed to represent the majority of the image saliency, the algorithm is applied iteratively until the majority is reached.

Figure~\ref{fig:final} displays an example of such color adaptation. 
The image is adapted for a method engineer, the position of the crane and its access are a crucial information for the construction site. 
Other elements represent structure components such as concrete walls, metal beams and frames. 
On the original image, the crane is red and the structure is orange, on the modified image the structure is green, the opposite colors of the crane.


\section{Experimental Validation} 
\label{sec:expe}

\subsection{Methodology and Experiments}

An experiment in real condition was setup, based on three real BIM models of Bouygues Construction projects from France, in the cities of Paris, Nantes and Lyon. 
Two of them, Paris and Nantes, mix structure elements (wall, roof, beam, ...) and construction methods elements (crane, scaffold, formwork). 
The third one, Lyon, includes plumbing elements (pipes). 
The three models were created on 3D building design software ``Revit'' and then exported in the real time 3D engine Unity3D\textregistered. 
Each element got a name and information about their color or kind of materials (concrete, metal, cloth...). 
In Unity3D\textregistered, these elements are automatically classified as structure, method or plumbing elements using the building ontology developed for the system.

Four visual scenes were taken for each model, and their associated depth maps were computed. 
Our system determines the color modification for each user profiles available with the 3D model. 
As the Paris and the Nantes models mix structural and method elements, four images with enhanced colors for structure engineers and four images with enhanced colors for method engineer were created. 
For the Lyon model, the same process is applied to obtain four images. 
There is a total of 20 enhanced color pictures.

\begin{figure}[ht]
\centering
\includegraphics[width=1\linewidth]{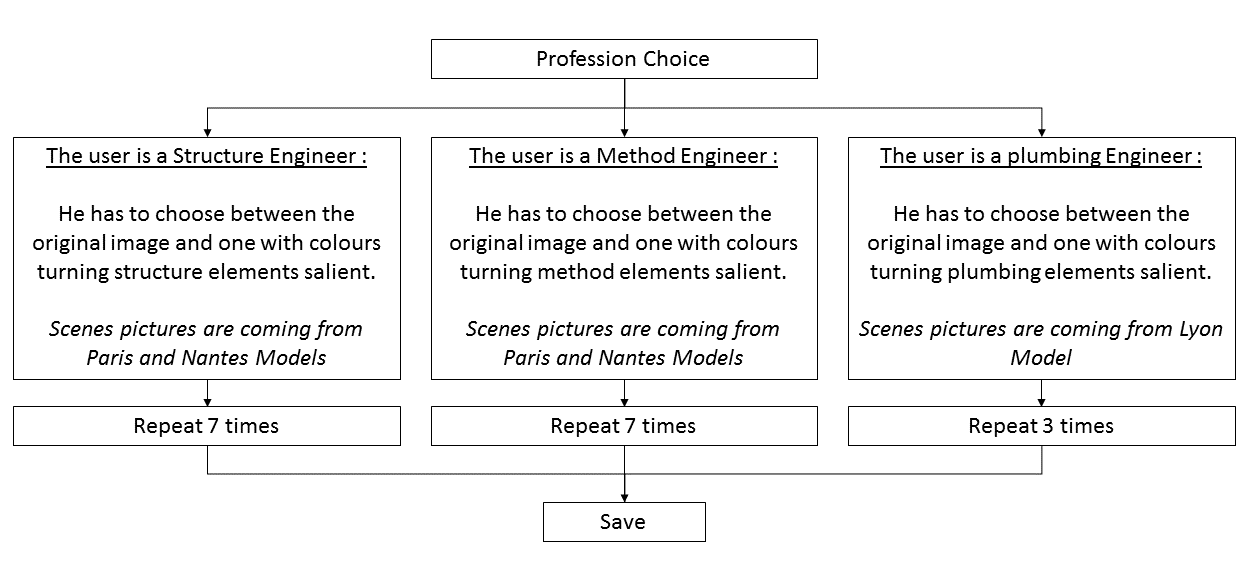}
\caption{Test protocol for the assessment of the proposed system.}
\label{fig:protocole}
\end{figure}

\begin{figure}[ht]
\centering
\includegraphics[width=1\linewidth]{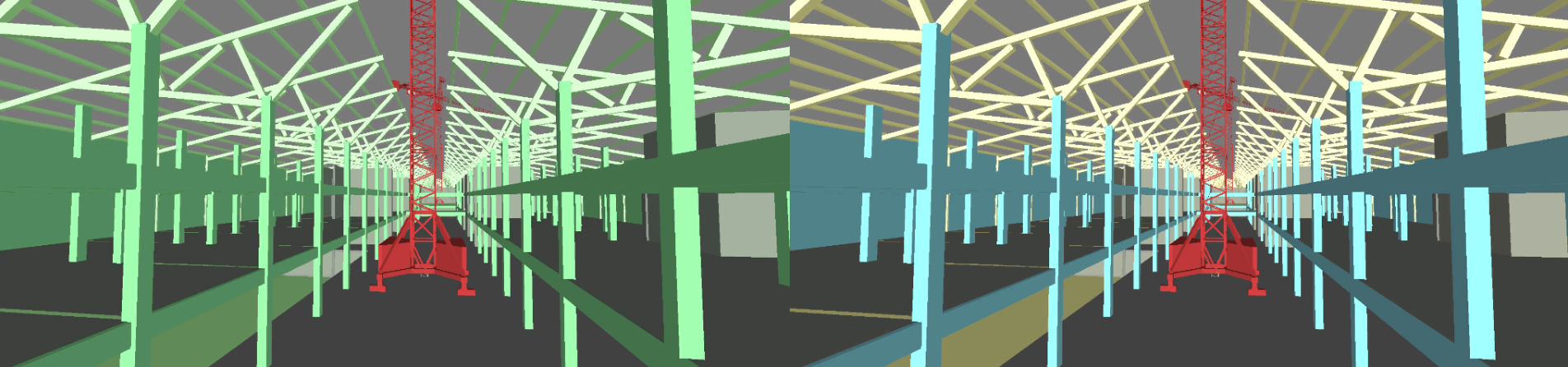}
\caption{Example of an image of the experiment. Experts are asked to answered the following question: "Considering that your are mostly interested by your profession's elements. In your opinion, which picture seems the more visually pleasant?" Five choices are allowed for the experts: ``left'', ``probably left'', ``indifferent'', ``probably right'' or ``right''. The right image is the original one; on the left one, the colors are enhanced by our saliency processing. The luminance are the same, only the color coded information are changed to emphasize the method objects (here the red crane in the middle). The blue and yellow elements from original image (right) are changed to green in the saliency-enhanced image (left).}
\label{fig:question}
\end{figure}

The experiments aim at evaluating the benefits of our method with building experts. 
A questionnaire was proposed to construction engineers, as illustrated on Fig.~\ref{fig:protocole}. 
In the experiments, the user profile (structure, method or plumbing engineer) is recorded.
Two images of the same visual scene are displayed on screen, chosen randomly from the 3D models: one image is the original, the other one is the color enhanced version.
The position, left or right, of the original and modified image is also chosen randomly for each presentation.
The expert is then asked to choose which image is the most confortable to work with.
The questionnaire have 5 possible choices: left image, probably left image, indifferent, probably right image and right image.
The response time for each image is also recorded.
An example of the setup is shown on Fig.~\ref{fig:question}. 
The structure and method experts are tested with 8 images extracted from the Paris and Nantes models, while the plumbing experts are tested on 4 images issued form the Lyon model.
A total of 17 experts have participated to these experiments: 6 are  method engineers, 7 are structure engineers and 4 are plumbing engineers.


\subsection{Results}

\begin{figure}[ht]
\centering
\includegraphics[width=0.6\linewidth]{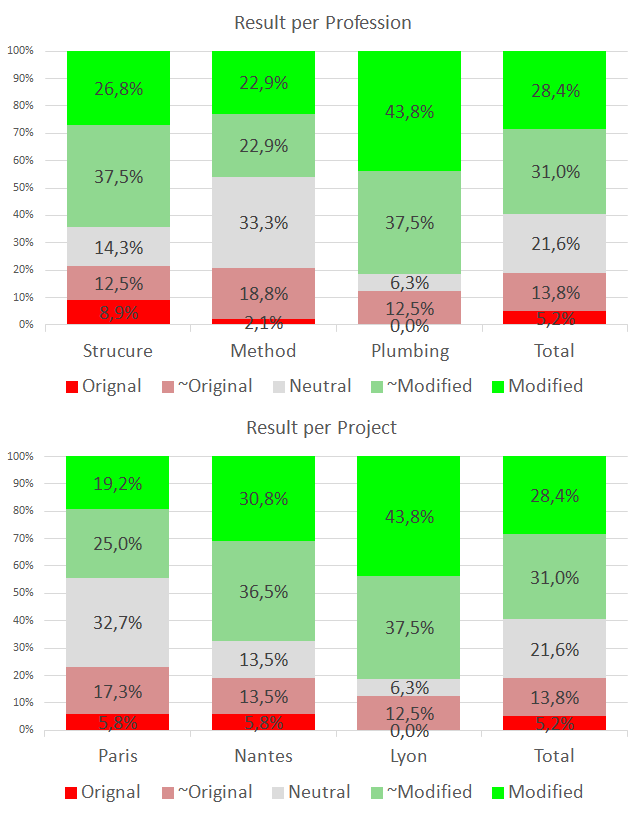}
\caption{Histograms of experts' answers. Bottom: Results for Paris, Nantes, Lyon. Top: Results for structure engineers, method engineers and plumbing engineers.}
\label{fig:graph}
\end{figure}

The results obtained from the experiments are shown on Fig.~\ref{fig:graph}.
The results are consistent across projects and user profiles. They do not indicate a bias for specific model or for a specific profession.
In average, the saliency-enhanced image is chosen in 59.4\% of all cases, to be compared with the 19\% of expert response selecting the original image.
The results confirm that our system improved the visualization quality while maintaining all information on screen.
Nonetheless, the ``indifferent'' answers indicate that there is room to improve the sytem.
The experts have confirmed that the have seen the advantage of the system and are waiting for the industrial BIM viewer version.

We applied Wilcoxon's Signed-Rank test. For each picture, we count the number of engineers who chose the modified or probably modified for the first sample, and those who chose probably the original or the original for the second sample. For the p-value decided to take 0.05 as limit, the p-value result is 0.001 which confirm the signifiance of the experimentation.


A limitation appeared during the experiments: the method experts inspecting the Paris model have shown some trouble deciding which image is the more comfortable.
This model contains blue/green scaffolds and security fences were red which is the opposite color. 
None of the two colors is dominant so the saliency-enhanced picture is not so different from the original. 
This problem is caused by elements of opposite colors belonging to the same category, due to the fact that the original elements colors are chosen randomly.
This limitation could simply be addressed by modify the original colors prior to the processing.

\section{Conclusion and perspectives}
\label{sec:ccl}

This paper describes a system for enhancing the visualization of BIM model, using saliencies to bias the visual focalization on the most relevant elements, given a specific user profile.
From the knowledge of BIM managers, we proposed a classifier to sort elements of the model depending of their category regarding per building departments.
We described an adapted saliency system to architectural images which implements a perspectivity integration for the orientations map and integrated depth clues.
Finally the system will be used in an application for the assistance for BIM engineers.
The system adapts model colors to keep the visual attention of the important elements for the user profile.
An experimentation was realized and show positively significant results and perspectives for the global improvement.

Future works will be dedicated to extending the system with other type of transformation for improving the visualization. 
Possible candidates includes transparency and spatial position.
It should be noted that the building ontology embedded in this system allows to specifies which transformation are allowed and which are not.
The ontology could be improved, to include elements defined as crucial for several departments.
Also, the ontology could include some priority information for the visualization of complex objects.
The last improvement concerns the position and orientation of the virtual camera (the user's view), which can help to define more appropriate enhancement.



\section*{Acknowledgments}
This work  was made with the collaboration of Bouygues Construction.
The resulting application will be integrated in the future BIM immersive room.
The research has been financed in partnership with the ANRT (Association Nationale de Recherche et Technologie).

\section*{References}

\bibliographystyle{plainnat}
\bibliography{bibBIM}

\end{document}